\documentclass[conference]{IEEEtran}
\IEEEoverridecommandlockouts
\usepackage{cite}
\usepackage{authblk}
\usepackage{amsmath,amssymb,amsfonts}
\usepackage[titlenumbered, ruled, linesnumbered, vlined]{algorithm2e}
\usepackage{graphicx}
\usepackage{caption,subcaption}
\usepackage{textcomp}
\usepackage{xcolor, colortbl}
\usepackage{todonotes,soul}
\usepackage[a4paper, total={184mm,239mm}]{geometry}
\usepackage{multirow}
\def\BibTeX{{\rm B\kern-.05em{\sc i\kern-.025em b}\kern-.08em
    T\kern-.1667em\lower.7ex\hbox{E}\kern-.125emX}}
\usepackage{amsmath,amssymb}
\let\oldnl\nl
\newcommand{\nonl}{\renewcommand{\nl}{\let\nl\oldnl}}
\def\fixme#1{\typeout{FIXED in page \thepage : {#1}}
\bgroup \color{red}{[FIXME: {#1}]} \egroup}

\begin{document}

\title{Anytime-Lidar: Deadline-aware 3D Object Detection}
\author[1]{Ahmet Soyyigit}
\author[2]{Shuochao Yao}
\author[3]{Heechul Yun}
\affil[1,3]{University of Kansas, Lawrence, KS\\ \{ahmet.soyyigit,heechul.yun\}@ku.edu}
\affil[2]{George Mason University, Fairfax, VA\\ shuochao@gmu.edu}

\maketitle

\begin{abstract}

In this work, we present a novel scheduling framework enabling anytime perception for deep neural network (DNN)
based 3D object detection pipelines. 
We focus on computationally expensive region proposal network (RPN) and per-category multi-head detector components, which are common in 3D object detection pipelines, and make them deadline-aware. We propose a scheduling algorithm, which intelligently selects the subset of the components to make effective time and accuracy trade-off on the fly. We minimize accuracy loss of skipping some of the neural network sub-components by projecting previously detected objects onto the current scene through estimations. We apply our approach to a state-of-art 3D object detection network, PointPillars, and evaluate its performance on Jetson Xavier AGX using nuScenes dataset. Compared to the baselines, our approach significantly improve the network's accuracy under various deadline constraints. 

\end{abstract}


\begin{figure*}[htbp]
\centerline{\includegraphics[scale=0.5]{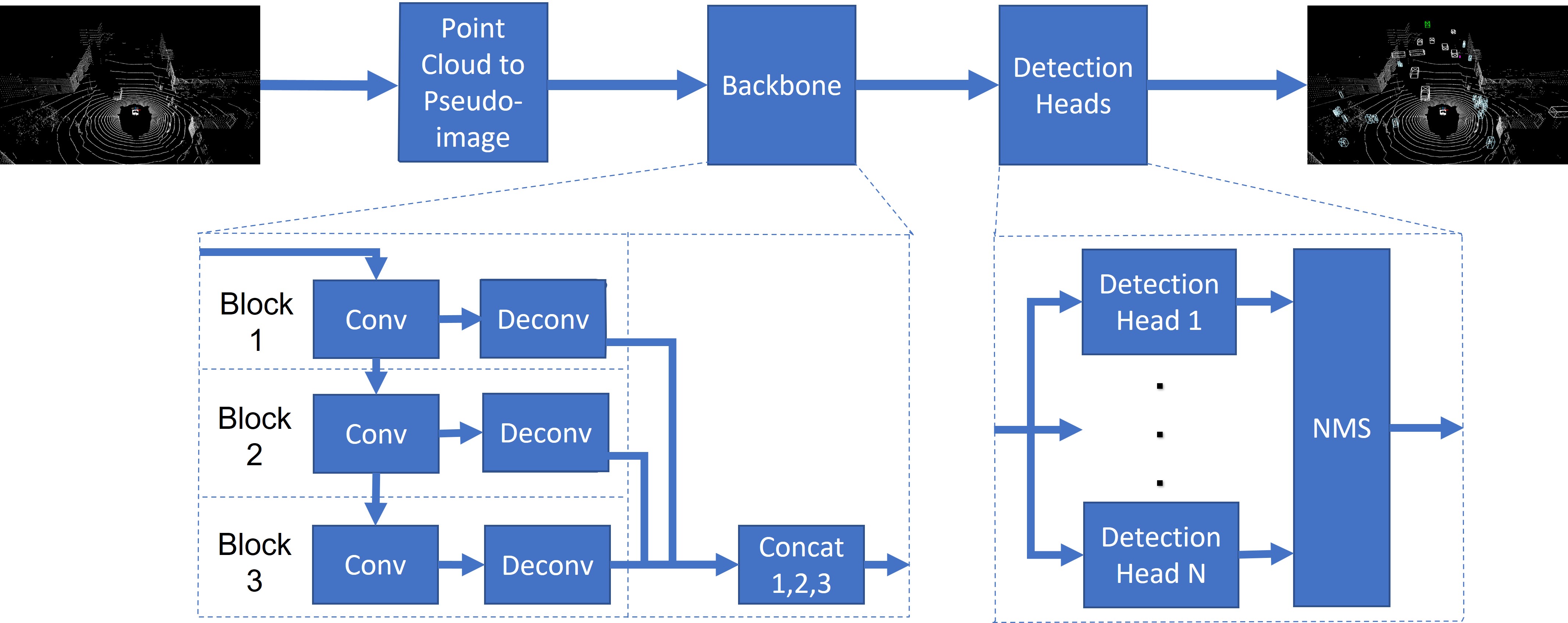}}
\caption{Baseline lidar detector architecture.}
\label{baseline}
\end{figure*}

\section{Introduction}\label{sec-intro}

Real-time object detection is a crucial part of autonomous vehicles. Radars, cameras and lidars are commonly utilized to provide sensory input for this task.
All detection systems aim to detect objects accurately and timely in order to make local and global path planning safe and efficient. 

Deep learning-based object detection systems have gained popularity in recent years due to their excellent performance. As a result, most object detection systems in autonomous vehicles, as in many other AI-related fields, are deep learning-based, which utilize neural networks.
These networks are composed of directed acyclic predefined tensor operations, executed layer by layer to obtain the detection results. Therefore, their computational cost and execution time is deterministic and highly predictable. While this predictability in itself could be a good virtue in a real-time system, it also means that the object detection system's timing and computational properties cannot be changed dynamically.

Within an autonomous driving framework, however, the timing (deadlines) and performance (accuracy) requirements of the detection tasks can change dynamically over time. For instance, a fast moving vehicle on a highway may favor less accurate but faster detection system over highly accurate but slow one. On the other hand, a slow moving vehicle on a complex city driving condition may favor higher accuracy even if it takes longer to detect the objects~\cite{multipath}. 
Also, hazardous situations such as sudden jaywalker appearance may need to trigger path planning, which may require reallocation of computing resources to process more urgent tasks in time. 
Therefore, it is crucially important to be able to make time and accuracy trade-off on the fly so the system can adapt to dynamically changing environment and make best use of the on-board computing resources, which are limited due to size, weight, power and cost constraints in autonomous vehicles.  

There have been growing number of research effort to support deadline-aware anytime perception capabilities in the real-time community. Several researchers proposed to apply anytime computing in vision based image classification by dynamically skipping some layers\cite{multipath,anytimenet,apnet,imprecisecomp} or prioritizing subset of neurons~\cite{subflow} in the backbone networks, while others applied similar ideas to vision based real-time object detection problems~\cite{multipath,rttasksched}.
Our work extends the anytime perception problem to 3D object detection utilizing lidar point clouds.  

The majority of today's cutting-edge neural network architectures can be broadly viewed as an encoder-decoder structure~\cite{voxelrcnn,fasterrcnn,pointpillars,yang2019std,zhou2020tracking}. The encoder (i.e., backbone network) extracts features from the input data, while the decoder converts those features into application-specific outputs.  
Although the backbone network of an image-based neural network is much more complicated than the decoder, the lidar-based backbone module often performs similar or less computation than the decoder.
Thus, it is necessary to address significant discrepancies between vision-based and lidar-based detection tasks for anytime inference.
On the one hand, this reduces the effectiveness of dynamic layer skipping and subnet prioritization exclusively in the backbone, since the decoder becomes another computational bottleneck. On the other hand, in contrast to the vision-based anytime model, the lidar-based network employs novel architectures such as multiple-scaled backbone blocks~\cite{voxelrcnn,fasterrcnn,pointpillars} and multiple detection heads~\cite{pointpillars,yang2019std,zhou2020tracking} to improve detection accuracy over a wide range of object sizes and categories, offering a new research challenge for enabling anytime predictions.

To this end, we propose a novel scheduling framework that enables anytime perception for lidar-based 3D neural object detection pipelines. 
To begin, we enable imprecise computing on the encoder of the neural network (i.e. backbone). We focus on multi-scaled Region Proposal Networks (RPNs),
which are widely employed in lidar and vision-based object detectors~\cite{fasterrcnn,voxelrcnn,pointpillars,redmon2018yolov3} but have received little attention in the prior anytime computing literature.
To our knowledge, the only exception~\cite{multipath} executes the whole RPN module but dynamically reduces the amount of output proposals in order to lower the cost of subsequent processing. However, they still fail to perform imprecise computation on the RPN module itself.
The multi-scaled RPN is structured as a series of blocks each of which generates bounding box proposals with increasing spatial resolutions. We exploit this structural characteristics and dynamically adjust the number of blocks to be executed at runtime in order to achieve a trade-off between speed and accuracy. 

Secondly, we further apply imprecise computation to the decoder of neural network (i.e., detection heads). 
Given that a lot of state-of-the-art lidar-based object detection networks utilize multiple detection heads~\cite{pointpillars,yang2019std,zhou2020tracking}, we present a head-skipping mechanism, that allows for the selective execution of a portion of the detection heads at runtime to further optimize the time and accuracy trade-off.

Thirdly, we introduce a lightweight projection method to compensate for the skipped detection heads in the current lidar frame. Assuming a fine-grained localization algorithm is providing global position of the ego-vehicle at runtime, this method allows estimating the detection results of the skipped heads by using the past detection results, improving the detection performance while avoiding potential dangers associated with skipped detection heads.

Lastly, we propose a two-phase dynamic scheduler, which decides the execution patterns of imprecise-computation capable encoder and decoder of the network prior to their execution. The scheduler is designed to maximize the detection accuracy while considering the deadline constrains. 


As a case study, we apply our technique on state-of-the-art lidar object detector, PointPillars\cite{pointpillars}. We evaluate our technique on a commercial-off-the-shelf embedded computing platform, Jetson Xavier AGX \cite{jetson-agx}. Our evaluation includes comparison with a range of methods including alternative scheduling methods that are capable of utilizing imprecise computation capable encoder and decoder. The results show that Anytime-Lidar can reduce its runtime requirement by 50 percent while meeting all the deadlines and providing better accuracy compared to baseline, multi-path, and alternative scheduling methods.

To the best of our knowledge, this work is the first to tackle the anytime perception problem in the lidar domain by considering the differences of 3D object detection compared to 2D. Our scheduling framework addresses the specific needs of 3D object detection to enable fine-grained time and accuracy trade-off.


\section{Background}\label{sec:background}

\begin{figure*}[htbp]
\centerline{\includegraphics[scale=0.5]{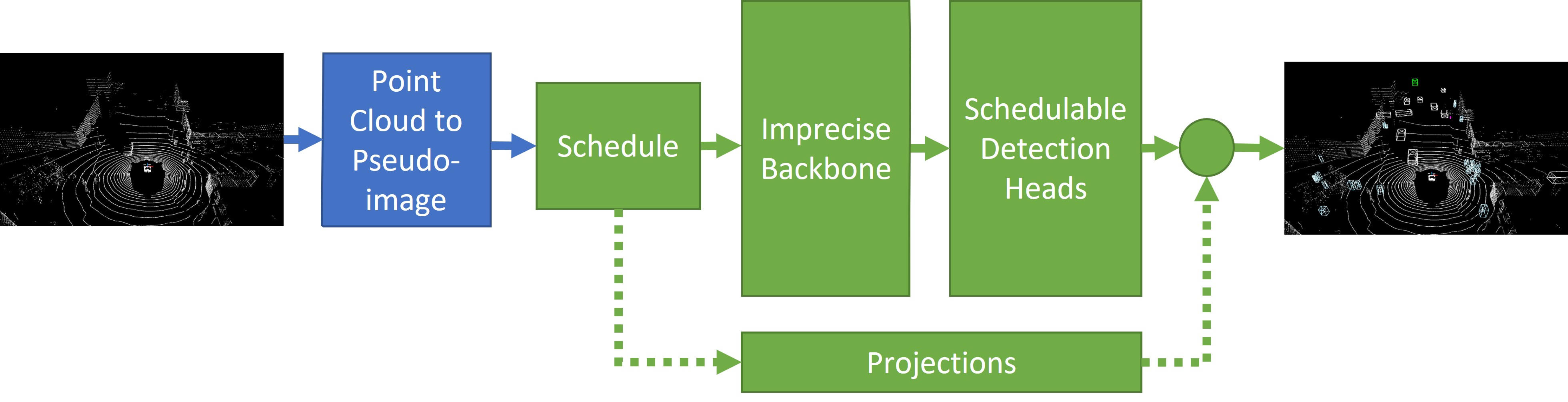}}
\caption{Proposed anytime lidar detector architecture.}
\label{proposed}
\end{figure*}

The main goal of object detection using point clouds is to produce the 3D positions, bounding boxes, and optionally velocities of the objects present on the scene. Figure~\ref{baseline} shows the general flow of the deep learning based techniques, which is our focus in this paper. Firstly, we see the initial stage transforming the point cloud into a pseudo-image of extracted features at runtime so the backbone, which is usually a convolutional neural network, can process it. This is required as a fact that unlike images, point clouds are not represented in a 2D array of pixels that can be directly processed with CNNs. Different techniques of transformation exist while all having the goal of producing a bird-eye-view representation of the scene. 


Afterwards, the generated bird-eye-view pseudo-image is processed by the backbone, which is usually Region Proposal Network (RPN) \cite{fasterrcnn} as can be seen in \cite{pvrcnn,second,centerpoint,parta2,cadnn,voxelrcnn,pointpillars}. RPN consists of multiple processing blocks, generally two or three, depending on the design choice. Figure~\ref{baseline} shows the inner structure of RPN with three blocks as in \cite{pointpillars}. Each block consists of a set of convolutions followed by a deconvolution operation that does upsampling. At the end, the output of all blocks are concatenated to produce a tensor of region proposals about where the objects are expected to be on the scene.

Lastly, the tensor of region proposals is processed by the detection heads. Each detection head is dedicated to derive the prediction results for a subset of the object classes through a set of convolutions. These predictions include the classification scores, positions, bounding boxes, and velocities of the objects. Non-maximum suppression is applied individually to the output of each detection head to obtain the final results.

It is possible to use a single detection head to produce the output for all classes, as it is the norm in image processing. When detection is happening in 2D, the bounding box size for a specific class can have highly varying size depending on the position of the object. For example, a bus within a short distance can be captured with a bounding box that is four times larger than the bounding box of a bus far away from the ego-vehicle. When the detection is performed in 3D, however, the bounding boxes become 3D and their size stays consistent regardless of the positions of the objects. This results the classes to have a bonding relationship with their bounding box anchor sizes, in contrast to image processing. In order to exploit this fact, the classes are divided into subsets according to their anchor sizes and an individual detection head is dedicated for each of these subsets\cite{megvii}. For example, one detection head is responsible for detecting pedestrians and traffic cones while another one does the same for buses and trucks. The utilization of multiple detection heads improves the detection accuracy by allowing each detection head to focus on a subset of classes having similar size. The downside is the increased overhead of running multiple detection heads instead of a single one, which has to be tackled for real-time execution.

\section{Anytime-Lidar}\label{sec:anytimelidar}
In this section, we present our method of making fine-grained time and accuracy trade-off at runtime for lidar-based object detectors. Our goal is to allow meeting tight deadlines which baseline models fails to do while maintaining higher detection accuracy for all deadline cases.

In this work, we base our method on PointPillars\cite{pointpillars}, one of the state-of-the-art detectors capable of delivering strong detection results with minimal latency, compared to other state-of-the-art methods. Our method does not involve modifying any part that is specific to PointPillars, therefore, it can be applicable to many other object detection architectures that utilize Region Proposal Network as backbone and multiple detection heads, such as \cite{second,centerpoint,pixor,megvii}.

\subsection{Overview}
The general architecture of the proposed method is shown in Figure~\ref{proposed}. 
We focus on enhancing the backbone and the detection heads to be flexible in their timing as they are the main source of inference latency. 
Table~\ref{exectimetable} shows the execution time profiling data of the baseline PointPillar network on Jetson AGX Xavier. 
It can be seen that the processing of backbone and detection heads account for 80\% of the total execution time on average. 

In a nutshell, our enhancements can be described in four parts. First, we allow backbone to run varying number of blocks. Second, we enable skipping of a subset of detection heads in favor of others. Third, we provide output for the skipped heads by projecting the past detection results to the current frame. Lastly, the stage/head scheduler manages the other three modules and decide what number of blocks of the backbone and which detection heads are to be executed. 
In the following subsections, we will give the details of our enhancements.

\begin{table}[t]
\caption{Execution Timing Analysis of PointPillars on Jetson AGX Xavier.}
\begin{center}
\begin{tabular}{|l|c|c|c|c|}
\hline
Stage & Min & Average & 99Perc & Percentage\\
\hline
PC Transformation & 24.61 & 26.64 & 28.40 &  20\%\\
\hline
Backbone & 44.81 & 45.23 & 45.55 &  34\%\\
\hline
Detection Heads & 55.06 & 58.83 & 62.72 &  46\%\\
\hline
\multicolumn{5}{l}{* Numbers are in milliseconds except last column.}
\end{tabular}
\label{exectimetable}
\end{center}
\end{table}

\begin{figure}[t]
\centerline{\includegraphics[scale=0.5]{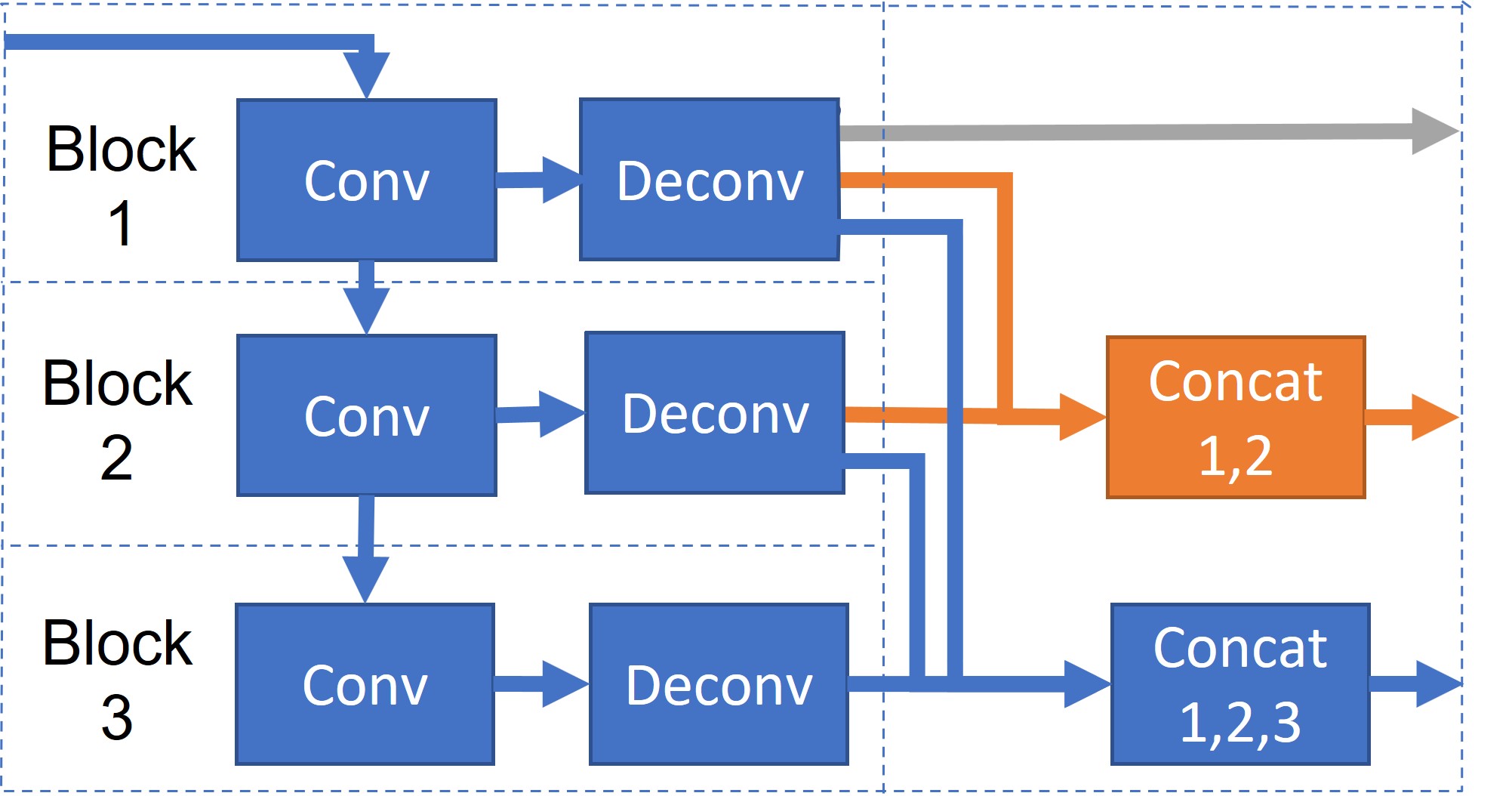}}
\caption{Imprecise backbone.}
\label{impbackbone}
\end{figure}

\subsection{Imprecise computation on backbone} \label{sec:imprecise-backbone}

We take advantage of the RPN's multi-scale block structure and add two early exists to the end of first two blocks as shown in Figure~\ref{impbackbone} to allow time and accuracy trade-off. This minimal change does not involve adding any additional network layer while allowing it to run reduced number of blocks at runtime, depending on the scheduling decision. 

It should be noted that connecting all the exit points of the imprecise backbone to the same set of detection heads would complicate the training process as the detection heads won't be able to focus on the output of a specific exit. As a solution, we duplicate the detection heads for each exit and train the entire network once with the method proposed in \cite{abc}, which sets training loss weights for each exit of the RPN that change over epochs. This enables each detection head to be trained to focus an exit of the backbone to achieve the best detection accuracy.

\begin{figure}[htbp]
\centerline{\includegraphics[scale=0.5]{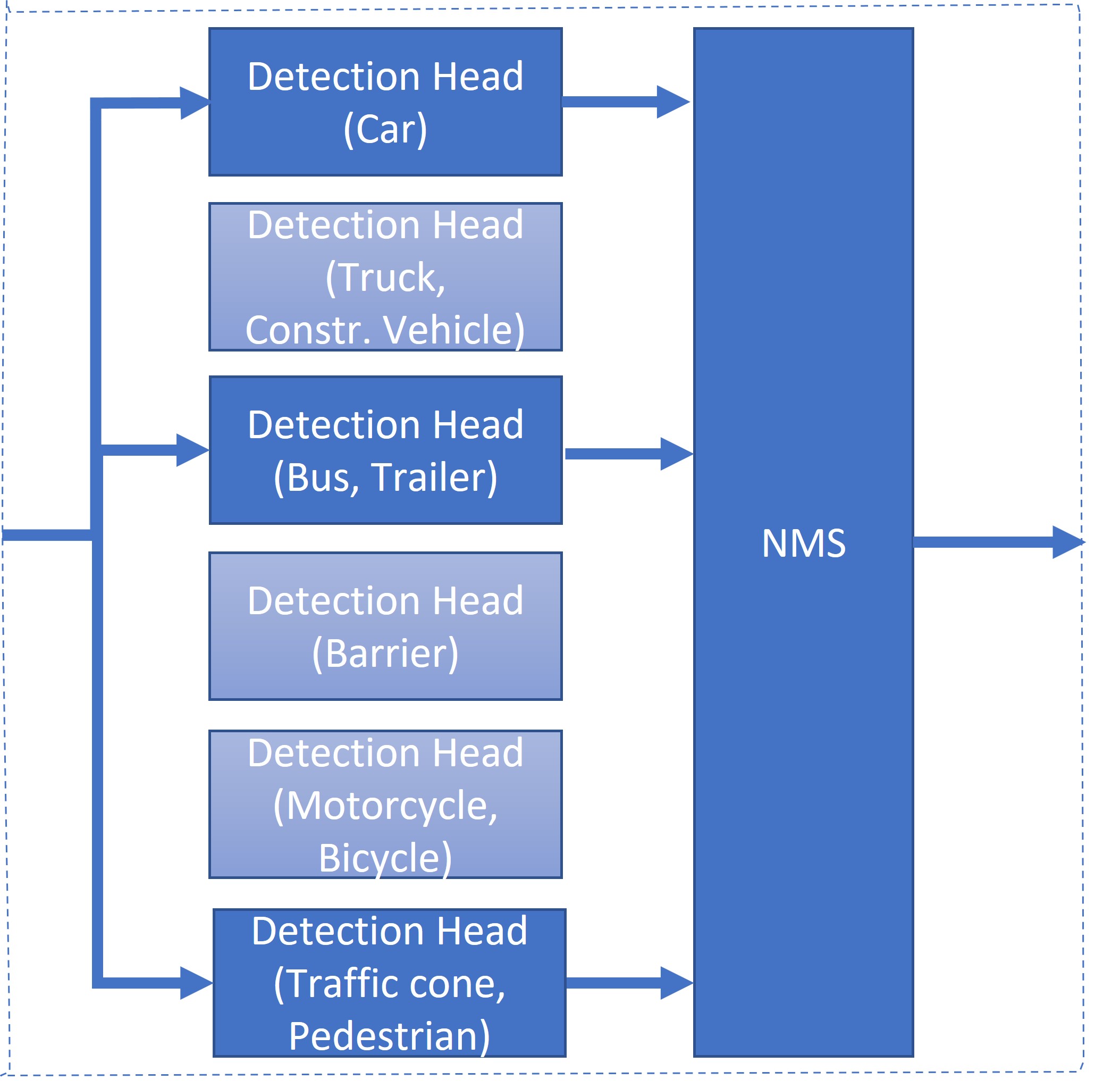}}
\caption{Schedulable detection heads.}
\label{detheads}
\end{figure}

\subsection{Detection head skipping}

To enable fine-grained execution time flexibility, we allow skipping of detection heads as illustrated in Figure ~\ref{detheads}. Each skipped head saves us from running a set of convolutions dedicated to output the position, size, classification score, and velocity of the objects. In addition, skipping a head also saves us from running non-maximum suppression on its output.

The reader might be concerned about the impact of head skipping on safety and detection accuracy. When heads are skipped, any object belonging to the subset of classes dedicated to the skipped heads won't be detected at all. However, this is only the case for single a lidar frame. Not all objects on the scene have to detected in every single frame but they should be detected within a range of consecutive frames. This can be ensured by scheduling the heads in a way that every head get its turn within a time limit. We handle this problem by our head scheduling method, which will be explained later. Also, the projection module mitigates the possible negative effect of skipped heads.

\subsection{Projection}
\label{sec:projection}

We integrate a projection module to the proposed method that estimates the current position, orientation, and velocity of the previously detected objects and use it to provide output in place of the skipped detection heads. This estimation is based on the assumption that the global position of the ego-vehicle, i.e. ego-pose, is available for each frame. In a typical autonomous driving framework, a localization module calculates ego-poses periodically so the planning can be made. Therefore our assumption does not add an extra overhead to the overall system.

The detector outputs the bounding boxes in lidar coordinate system, which is attached to the ego-vehicle. This coordinate system changes as the vehicle moves. Therefore, projecting a bounding box requires applying 3D transformations to move it to the current lidar coordinate system from the past one. Also, the last known velocity of the object should be utilized to make projection accurate. Therefore, we follow the below steps to project a bounding box:
\begin{itemize}
    \item Rotate and translate the object to be in the ego-pose coordinate system of the time of detection.
    \item Rotate and translate the object to be in the global coordinate system.
    \item Translate the object using its last known global velocity and the elapsed time since it was detected.
    \item Translate and rotate the object to be in the current ego-pose coordinate system.
    \item Translate and rotate the object to be in the current lidar coordinate system.
\end{itemize}

The listed operations has to be performed for all bounding boxes to be projected, which can introduce significant computational overhead. We tackle this problem by taking advantage of the available CPU resources while the neural networks execute on the GPU. After we determine all the bounding boxes to be projected, we distribute them to parallel CPU processes. While they are being processed on the CPU cores, we continue with the execution of backbone and detection heads on the GPU. When the detection heads finish running, we collect the projected bounding boxes and merge them with the detection results.

\begin{table}[t]
\caption{Post-sync WCET calibration table for Jetson AGX Xavier}
\begin{center}
\begin{tabular}{|l|c|c|c|c|c|c|}
\hline
RPN & \multicolumn{6}{c|}{Detection heads}\\
\cline{2-7}
blocks & 1 & 2  & 3 & 4 & 5 & 6\\
\hline
1 & \cellcolor{green!15}30.9 & \cellcolor{green!15}42.2 & \cellcolor{green!15}52.2 & 62.1 & 70.6 & 78.2\\
\hline
2 & \cellcolor{green!15}46.3 & \cellcolor{green!15}56.8 & \cellcolor{green!15}66.9 & \cellcolor{green!15}76.8 & \cellcolor{green!15}85.4 & 93.2\\
\hline
3 & \cellcolor{green!15}61.8 & \cellcolor{green!15}71.9 & \cellcolor{green!15}81.8 & \cellcolor{green!15}92.0 & \cellcolor{green!15}100.6 & \cellcolor{green!15}107.9 \\
\hline
\multicolumn{7}{l}{* Numbers are in milliseconds.}
\end{tabular}
\label{calibrationWCET}
\end{center}
\end{table}

\begin{table}[t]
\caption{Normalized accuracy calibration table (\%)}
\begin{center}
\begin{tabular}{|l|c|c|c|c|c|c|}
\hline
RPN & \multicolumn{6}{c|}{Detection heads}\\
\cline{2-7}
blocks & 1 & 2 & 3 & 4 & 5 & 6\\
\hline
1 & \cellcolor{green!15}{67.0} & \cellcolor{green!15}{67.5} & \cellcolor{green!15}70.7 & 74.4 & 79.2 & 80.6\\
\hline
2 & \cellcolor{green!15}{75.4} & \cellcolor{green!15}{77.5} & \cellcolor{green!15}{82.1} & \cellcolor{green!15}{88.2} & \cellcolor{green!15}{91.9} & 93.3\\
\hline
3 & \cellcolor{green!15}{79.8} & \cellcolor{green!15}{84.9} & \cellcolor{green!15}{90.7} & \cellcolor{green!15}{95.6} & \cellcolor{green!15}{98.9} & \cellcolor{green!15}{100.0}\\
\hline
\end{tabular}
\label{calibrationAcc}
\end{center}
\end{table}

\subsection{Scheduling of RPN blocks and detection heads}

The scheduling problem involves deciding the number of backbone blocks and which detection heads to run. A certain deadline can be met by multiple block/head configurations, but how do we know which one would deliver the best detection accuracy? We solve this problem by dividing the scheduling into two phases. First, we determine the number of RPN blocks and detection heads to execute by using offline profiling data, represented in the form of two calibration tables, one for WCET and the other for accuracy. Second, we determine which detection heads to skip by our novel  head scheduling algorithm.

The calibration tables hold the required execution time and deliverable detection accuracy for all combinations of number of RPN blocks and number of detection heads. Using a small subset of the training data, we conduct the calibration with the procedure shown in Algorithm~\ref{alg:algo1} to reveal how accuracy and execution time changes depending on the block/head configuration. Once the calibration is done, the produced tables are saved in disk and loaded to the memory whenever a test is conducted. Tables~\ref{calibrationWCET} and~\ref{calibrationAcc} are the ones generated for our method. Please note that the Table~\ref{calibrationWCET} has the required execution times to complete the execution after point cloud transformed to pseudo-image. In the tables, the green cells are the executable block/head configurations. Other configurations deliver less or equal accuracy with higher demand of time, so they are not considered for execution.

\setlength{\textfloatsep}{2pt}
\begin{algorithm}[t]
\DontPrintSemicolon
\textbf{Input}:\\
Calibration dataset ($D$),\\
Number of RPN blocks ($R$),\\
Number of detection heads ($H$),\\
\textbf{Output}: WCET and Accuracy Tables\\
\SetKwProg{funct}{function}{}{}
\SetKwFunction{func}{calibrate}

\funct{\func{$D$, $R$, $H$}}
{
    $wcet\_table \gets 2D\_array\_with\_size(R,H)$\\
    $acc\_table \gets 2D\_array\_with\_size(R,H)$\\
    
    $r \gets 1$\\
    $h \gets 1$\\
    \While{$r \leq R$}
    {
        \While{$h \leq H$}
        {
            $fix\_num\_blocks\_and\_heads(r, h)$\\
            $deadline \gets \infty$\\
            $w, a \gets process\_samples(D, deadline)$\\
            $wcet\_table[r,h] \gets w$\\
            $acc\_table[r,h] \gets a$\\
            $h \gets h + 1$\\
        }
        $r \gets r + 1$\\
    }
}
\Return{$wcet\_table, acc\_table$}
\caption{Calibration Procedure}
\label{alg:algo1}
\end{algorithm}

\subsubsection{First phase of scheduling}

Initially, we wait until the point cloud transformation finishes its execution on the GPU since the GPU operations are executed asynchronously. This synchronization is done to calculate the remaining time with precision.
Afterwards, we iterate over the calibration tables to find the block/head configuration that would provide highest possible detection accuracy while meeting the deadline with respect to the remaining time. Once it is found, we move to the second phase.

\subsubsection{Second phase of scheduling}

We propose to schedule the detection heads by dynamically assigning priorities based on two parameters: their age and aged confidence. The age of a head is defined as the number of frames passed since the last time it was scheduled to run. We keep track of the ages of all heads and update them whenever a frame is processed, resetting the age of scheduled heads and increasing the age of others.

Each detected object has a confidence score alongside its bounding box. The confidence scores of all objects detected by a certain detection head gives a clue of its contribution to the overall detection accuracy, which can be used for prioritization. The aged confidence of a detection head is defined as the sum of these confidence scores at the last time it was scheduled to run. Although the utility of aged confidence decreases over time, it still provides a useful hint about which heads would contribute most to the overall detection accuracy thanks to the temporal locality of the objects on the scene. Since the time between the frames are in the order of milliseconds, the objects on the scene does not change significantly over a sequence of consecutive frames. We take advantage of this fact to save ourselves from the overhead of calculating the confidence of all heads for every single frame.

Algorithm~\ref{alg:algo2} summarizes the proposed detection head scheduling algorithm. Firstly, we update the aged confidences of the heads scheduled in the previous frame by taking their sum and normalizing them (line 8,9). Afterwards, we create an empty max heap of tuples which makes comparison using the first element of the tuples (line 11). We calculate the priority of each head by multiplying its age with its aged confidence and push it to the max heap (line 16). We make sure that any head with an age beyond a predefined frame limit is given a special priority by multiplying it with the maximum possible aged confidence (line 14). Lastly, we pop heads from the max heap and put them in the schedule list until we reach the number of heads determined to run at the first scheduling phase (lines 21-24).

\setlength{\textfloatsep}{2pt}
\begin{algorithm}[t]
\DontPrintSemicolon
\textbf{Input}:\\
Heads Scheduled For the Previous Frame ($H_{prev}$),\\
Aged Head Confidences ($C$),\\
Head Ages ($A$),\\
Number of Heads To Run ($N$)\\
\textbf{Output}: List of heads to run\\
\SetKwProg{funct}{function}{}{}
\SetKwFunction{func}{sched\_det\_heads}

\funct{\func{$H{prev}$, $C$, $A$, $N$}}
{
    \For{$h \in H_{prev}$}
    {
        $C[h] \gets norm(sum(get\_prev\_det\_scores(h)))$\\
    }

    $h \gets 1$ \\
    $prios \gets max\_heap()$\\
    \For{$c \in C$}
    {
        \If{$A[h] > $FRAME\_LIMIT}
        {
            $tuple \gets (A[h] \times $\emph{MAX\_SCORE}$), h)$
        }
        \Else
        {
            $tuple \gets (A[h] \times c, h)$
        }
        $prios.push(tuple)$\\
        $h \gets h + 1$\\
    }
    
    $i \gets 1$ \\
    $H \gets ()$ \\
    \While{$i \leq N$}
    {
        $scr, h \gets  prios.pop()$\\
        $H \gets H \cup h$\\
        $i \gets i + 1$\\
    }
    
}
\Return{$H$}
\caption{Head Selection Heuristic}
\label{alg:algo2}
\end{algorithm}

\begin{table*}[t]
\centering
\caption{Methods to compare}
\begin{center}
\begin{tabular}{|l|c|c|c|c|}
\hline
Method & Number of model  & Number of  & RPN stage & Detection head\\
       & parameters       & RPN blocks & selection & scheduling\\
\hline
PointPillars-3 & 6078K & 3 &  & \\
\hline
PointPillars-2 & 2626K & 2 &  & \\
\hline
PointPillars-1 & 1723K & 1 &  & \\
\hline
MultiStage & \multirow{6}{*}{9235K} & \multirow{6}{*}{3} & \multirow{6}{*}{\checkmark} & \\
\cline{1-1}\cline{5-5} 
RoundRobin &  & &  & Circulating\\
\cline{1-1}\cline{5-5} 
ClsScrSum &  &  &  & Class scores sum \\
\cline{1-1}\cline{5-5}
NearOptimal  & &  &  & Aging + Ground Truth \\
\cline{1-1}\cline{5-5} 
Ours   &  &  &  & Aging + Aged confidences \\
\hline
\end{tabular}
\label{methodsToCompare}
\end{center}
\end{table*}

\section{Evaluation}\label{sec:evaluation}
We evaluate the proposed method by extending the PyTorch implementation of PointPillars in \cite{openpcdet2020} to support our method and others we used for comparison. We present our method's performance under three contexts. Initially, we show the benefit of head scheduling plus projection over methods not having head-skipping capability. Next, the role of the head scheduling algorithm is highlighted. Last, the improvement that comes by the projection module is demonstrated.

We train the neural network models and test the methods using nuScenes \cite{nuscenes}, a realistic large-scale dataset for autonomous driving. nuScenes is composed of 1000 scenes where each scene has the data of a 20 seconds drive in urban environment. This data include the periodically collected samples of various sensors and annotations of the objects. In addition, periodically calculated global position of the ego-vehicle i.e. ego-poses are present in the dataset.

The evaluation procedure involves processing the lidar samples of five different 20 second scenes taken from the evaluation dataset. We execute this procedure for all methods within different deadline constraints and evaluate their performance in terms of detection accuracy and timeliness. We used nuScenes Detection Score (NDS)\cite{nuscenes} as our detection accuracy metric which includes in itself mean Average Precision as its parameter. We normalized the NDS scores for a better presentation of our results. Each deadline miss is counted as a detection with no results. Before running our tests, we do the required accuracy/WCET calibration using another five different scenes from the training dataset. Baseline methods don't require calibration as they run without scheduling.

One thing to note about the dataset is that the time between the consecutive annotated point clouds (i.e. keyframes) is 500 milliseconds whereas the point clouds were sampled with 50 milliseconds period. Since we can only evaluate the detection accuracy using the keyframes, the large time gap significantly reduces the projection performance. To overcome this problem and make the evaluation more realistic, we interpolated the annotations between the keyframes and allowed all point clouds to have annotations so they can be used in evaluation. For the tested deadlines within the ranges of 140-110 ms and 100-50 ms, we tested with the lidar samples having period of 150 ms and 100 ms, respectively.


An overview of the methods we tested is given at the Table ~\ref{methodsToCompare}. The baseline PointPillars methods utilize separately trained models whereas others use the same imprecise model. The reason why imprecise model has higher number of parameters is the duplicated detection heads that is connected to the early exits of imprecise backbone. All methods capable of head scheduling follow the processing stages shown in the Figure ~\ref{proposed}. The only difference between them is their head scheduling algorithm i.e. second phase of the scheduling. The initial scheduling phase utilizing the calibration table is same for all methods noting each method has its own calibration table.

As the testing platform, we use Jetson AGX Xavier\cite{jetson-agx} having 16GiBs of RAM. All hardware clocks were maximized during the tests. The neural network layers and non-maximum suppression were executed on GPU while the head/stage scheduling and projections were done on the CPU.

\begin{figure}[t]
\centerline{\includegraphics[scale=0.6]{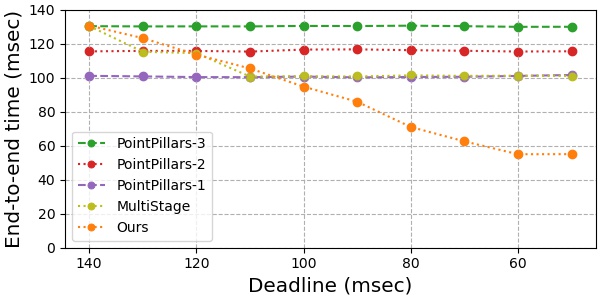}}
\caption{End-to-end average execution times.}
\label{ete1}
\end{figure}

\begin{figure}[t]

\centerline{\includegraphics[scale=0.6]{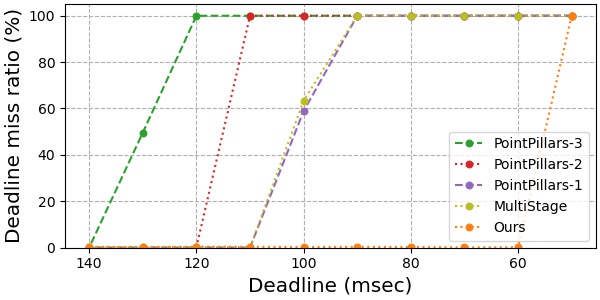}}
\caption{Deadline miss rates.}
\label{dl1}
\end{figure}

\subsection{Comparison With the Baselines and MultiStage}

Here, we highlight the main benefits of the proposed method which enable fine-grained execution time adjustment by comparing it with the following methods:

\begin{itemize}
\item \textbf{PointPillars}\cite{pointpillars}: These are the baseline models without any dynamic execution time adjustment capability. The numbers 1 to 3 denote the number of RPN blocks used in the backbone. Having less blocks reduces required execution time with a sacrifice from accuracy. Baseline-3 is the actual baseline model without any modification.
\item \textbf{MultiStage}\cite{multipath}: This method can make execution time/accuracy tradeoff by running different number of RPN blocks. However, it executes all detection heads for all frames.
\end{itemize}

\begin{figure}[t]
\centerline{\includegraphics[scale=0.6]{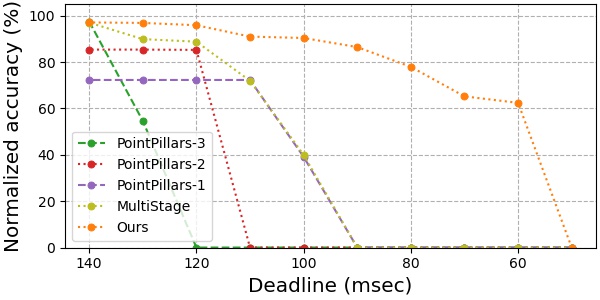}}
\caption{Detection accuracy versus baselines.}
\label{acc1}
\end{figure}

Figure ~\ref{ete1} shows how average execution time changes over deadlines. The baselines are rigid and has no capability of adjusting their execution timing, resulting deadline misses as shown in Figure ~\ref{dl1}. On the other hand, the proposed method can reduce its execution time requirement by reducing the number of executed backbone blocks and detection heads at runtime while providing higher accuracy than all baselines, as shown in Figure ~\ref{acc1}. The MultiStage method can also provide a degree of execution time flexibility and better accuracy results compared to baselines. However, its flexibility is limited with the backbone, consequentially falling behind the proposed method.

\begin{figure}[t]
\centerline{\includegraphics[scale=0.6]{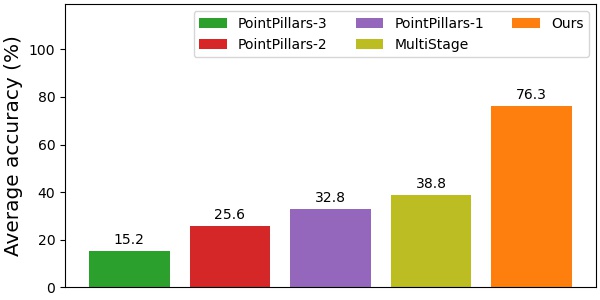}}
\caption{Average detection accuracy versus baselines.}
\label{acc1bar}
\end{figure}

The Figure ~\ref{acc1bar} shows the achieved average accuracy over all tested deadlines, it is basically the average form of the Figure ~\ref{acc1}. When we look the results in this figure side by side with the deadline misses, the main takeaway is that meeting the deadlines plays the most crucial role for achieving high overall detection accuracy. This can be seen even by comparing the first three baselines. The proposed method enables fine-grained execution time flexibility that allows meeting tight deadlines, which in turn boosts the average accuracy to be twice of what MultiStage can supply.

\subsection{Analyzing the impact of head scheduling and projection}

In this section, we investigate how head scheduling and projection, as separate components, impact the performance. For that purpose, we compare the proposed method with the alternative head scheduling methods listed below:
\begin{itemize}
\item \textbf{RoundRobin}: The detection heads are scheduled in a round-robin order, giving equal priority to all heads.
\item \textbf{ClsScrSum}: This method makes prioritization based on the sum of classification scores calculated on-the-fly. It runs the classification part of all detection heads and individually takes their sums after applying sigmoid function and filtering it with the predefined score threshold. The sums are multiplied with the age of each head similar to the proposed method to obtain final priorities. Afterwards, the remaining part of the chosen detection heads are executed while skipping others. 
\item \textbf{NearOptimal}: This method prioritizes the heads based on their ages, i.e. the time elapsed since their last usage, while avoiding to run any head that does not have a corresponding ground truth annotation on the frame being processed. Since this method relies on ground truth data, we use it as an upper bound of others to show the effectiveness of proposed method compared to this one.
\end{itemize}
\begin{figure}[t]
\centerline{\includegraphics[scale=0.6]{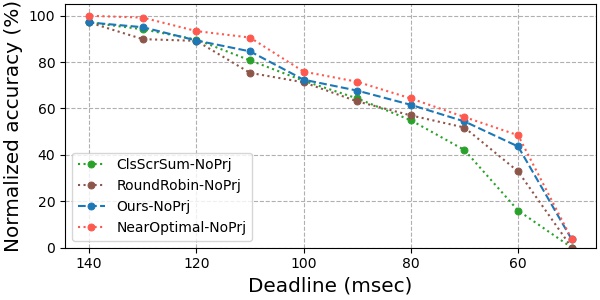}}
\caption{Detection accuracy of different head scheduling methods without projection.}
\label{acc2}
\end{figure}

\begin{figure}[t]
\centerline{\includegraphics[scale=0.6]{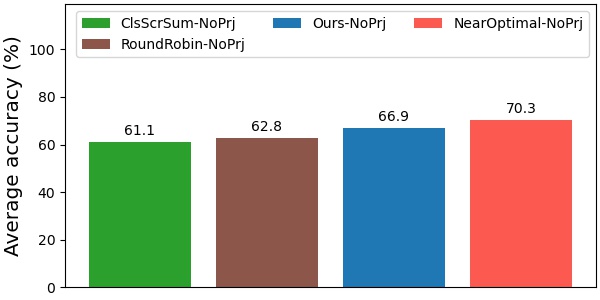}}
\caption{Average detection accuracy of different head scheduling methods without projection.}
\label{acc2bar}
\end{figure}

\subsubsection{Impact of head scheduling algorithm}

Figures ~\ref{acc2} and ~\ref{acc2bar} illustrates the detection performance of the aforementioned head scheduling algorithms when projection is disabled. RoundRobin circulates through all heads and provides a simple solution that maintains safety by letting each head to take its turn over time. However, it does not have a prioritization mechanism that can boost the accuracy by demoting the heads having low confidence. On the other hand, ClsScrSum and the proposed method utilize the confidence scores to prioritize the heads having high confidence, while maintaining the safety through aging. The overhead of ClsScrSum grows larger with the deadline being more tight, due to the requirement of running classification part of all heads at all times. The proposed method avoids this overhead by taking advantage of temporal locality, and gives the best results that is close to the near optimal solution.

\subsubsection{Impact of projection}

\begin{figure}[t]
\centerline{\includegraphics[scale=0.6]{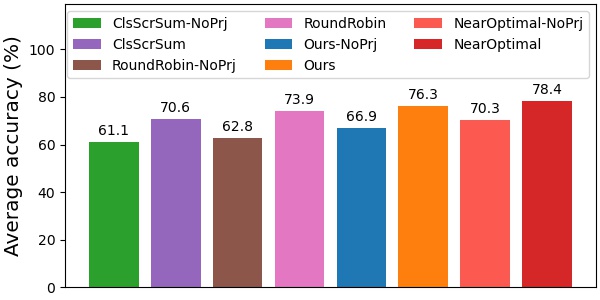}}
\caption{Average detection accuracy with and without projection.}
\label{acc3bar}
\end{figure}

Figure ~\ref{acc3bar} shows average accuracy improvement that comes by enabling projection of previously generated bounding boxes. Since projection is dependent on the recent history of detection head usage, all detection heads should be executed within a time limit so the projections can be made for the skipped heads. This can be ensured by either through round robin scheduling or aging. Because the compared four algorithms utilize either of these two, all of them enjoy a similar improvement through projection, as they are capable of keeping the projection data fresh.

\begin{figure}[t]
\centerline{\includegraphics[scale=0.6]{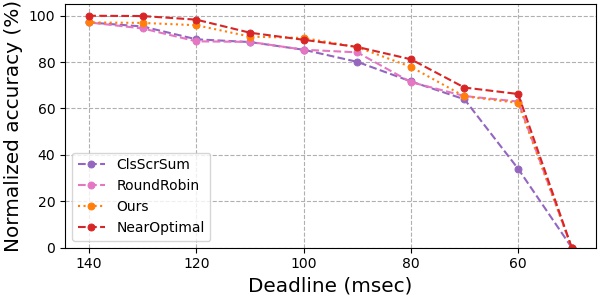}}
\caption{Detection accuracy versus other projection capable methods.}
\label{acc3}
\end{figure}

When we look at the Figure ~\ref{acc3}, we see that the proposed method is still the best performing one when it is compared to other projection enabled methods, and again it is very close to the near optimal solution. The takeaway is that the head scheduling algorithm still plays an important role when projection is enabled.

\begin{table}[t]
\caption{Average overhead.}
\begin{center}

\begin{tabular}{|l|c|c|c|}
\hline
Method & \multicolumn{3}{c|}{Overhead}\\
\cline{2-4}
             & Synchronization & Scheduling & Projection\\
\hline
RoundRobin   & 0.50            & 0.00       & 1.10\\ 
\hline
Ours         & 0.50            & 1.00       & 1.10\\
\hline
ClsScrSum    & 0.50            & 4.25       & 2.55\\
\hline
\multicolumn{4}{l}{* Numbers are in milliseconds.}

\end{tabular}
\label{overheadtable}
\end{center}
\end{table}

\begin{table}[t]
\caption{Average projection overhead for our method.}
\begin{center}

\begin{tabular}{|l|c|}
\hline
Projection technique & Overhead (ms)\\
\hline
Blocking single process   & 30.1\\ 
\hline
Blocking six processes    & 6.8\\
\hline
Asynchronous six processes    & 1.1\\
\hline
\multicolumn{2}{l}{* Numbers are in milliseconds.}

\end{tabular}
\label{overheadprojtable}
\end{center}
\end{table}

\subsection{Overhead Analysis}

Table ~\ref{overheadtable} shows the average additional time to finalize execution for the three versions of our anytime-lidar approach (differ only in head-selection method) compared to the baseline PointPillars network. Firstly, all listed methods suffer ~0.5 milliseconds overhead from the synchronization done prior to scheduling. Secondly, the detection head scheduling itself puts an additional overhead depending on the algorithm used. RoundRobin has the minimal overhead since the decision can be made in constant time, whereas the proposed method needs to calculation for prioritization. The main source of overhead for the ClsScrSum method is the requirement of running classification convolutions for all detection heads including the ones that will be scheduled to skip afterwards. Lastly, there is the overhead of projection, as the bounding boxes has to be distributed to the parallel processes, and at the end, the results should be collected and merged with the detection results. These two operations adds an additional overhead contingent on the method used. The ClsScrSum needs to do more projection as it skips more heads for any specific deadline compared to other methods. This is why it suffers most from the projection. In all, the total added overhead of our proposed approach (Ours) is approximately less than 3\% for a 100ms deadline.

We mentioned that projection overhead is mitigated with the use of asynchronous parallel processing in Section~\ref{sec:projection}. Table ~\ref{overheadprojtable} shows the alternative cases where the projections are done after the GPU execution finishes. The illustrated timings point out the significant mitigation of the projection overhead by asynchronous execution with parallel processes.

Our approach does increase the memory that model requires from 185 MiB to 281 MiB, as a result of the duplicated detection heads needed to support multi-exit imprecise backbone (Section~\ref{sec:imprecise-backbone}). Since the target platform has memory in the orders of GiB, we consider this  overhead acceptable.

\section{Related Work}
\label{sec:related}

Lidar based object detection is important for many autonomous driving frameworks~\cite{lidaro}.
The release of large-scale autonomous driving datasets ~\cite{nuscenes,waymo,lyft} allowed researchers to develop lidar-based deep learning techniques achieving remarkable detection performance. One of the state-of-the-art in this field is PointPillars ~\cite{pointpillars}, employed by industry and open-source autonomous driving frameworks ~\cite{autoware,apollo}. 
This method is nominated to deliver real-time performance for its high execution speed alongside strong detection accuracy. 
Besides PointPillars, many other methods have been presented promising low execution latency alongside strong detection accuracy~\cite{pixor,rtm3d,centerpoint,complexyolo,pvrcnn,parta2,second}. 
However, all these networks are still computationally expensive and cannot be dynamically adjusted in a deadline-aware manner.

In the broader AI community, many work has done on model compression such as weight quantization, pruning, etc.~\cite{deepcompression,deepiot,fastdeepiot,adaptivequant,cascaded}. 
Although these works are beneficial in reducing computational cost of execution neural networks, they do not support dynamic time and accuracy trade-off, which is needed for autonomous driving and other intelligent real-time systems.

Recently, many researchers explored anytime perception that enable deadline-aware neural network execution for real-time systems. For instance, Kim et al.~\cite{anytimenet} made time and accuracy trade-off possible for an image classification network by iteratively adding layers and re-training it to have early exits. Lee et al.~\cite{subflow} provided a solution in neuron level, prioritizing the subset of neurons which contribute most to the accuracy, while deactivating others to save time. Bateni et al.~\cite{apnet} used per-layer approximation instead of using early exits, and provided a scheduling solution for multiple DNN tasks. 
Yao et al.~\cite{imprecisecomp} also focused scheduling of multiple DNN tasks but using imprecise computation with early exits. In all, the focus of all these works was image classification whereas the proposed methods focus is object detection.
Heo et al.~\cite{multipath} proposed a multi-path DNN architecture for anytime perception in vision based object detection. 
Hu et al.\cite{hu2021exploring} proposed to reduce resolution of less critical part of the scene to reduce computational cost. 
Lie et al.~\cite{rttasksched, liu2020removing} break-up individual frames into smaller sub-regions with different criticality with the help of lidar and batch process important sub-regions to meet deadlines. 
However, all these effort mainly focus on 2D vision data processing and do not address the unique characteristics of 3D point cloud processing.

In contrast, our work address anytime perception problem in the lidar domain by considering the differences of 3D object detection compared to 2D.

\section{Conclusion}\label{sec:conclusion}

In this work, we presented a novel method of providing execution time flexibility to the machine learning based object detection pipelines utilizing point clouds as input. Results has shown that the proposed method delivers satisfactory detection accuracy over a long range of deadlines compared to baseline methods. This improvement comes from four major contributions, (i) modifying the backbone of the model to be partially executable by minimal effort, (ii) allowing a subset of detection heads to be skipped to enable fine-grained execution time and accuracy trade-off, (iii) integrating a projection mechanism that compensates potential accuracy loss due to skipped heads, (iv) and a two-phase scheduler that manages the execution of backbone, detection heads, and projection with the goal of maximizing detection accuracy while meeting the deadlines.
\section*{Acknowledgments}\label{sec:acknowledge}

This research is supported in part by NSF grants CNS1815959, CPS-2038923, and CPS-2038658.

\bibliography{references}

\begin{thebibliography}{10}

\bibitem{multipath}
S.~Heo, S.~Cho, Y.~Kim, and H.~Kim, ``Real-time object detection system with
  multi-path neural networks,'' in {\em 2020 IEEE Real-Time and Embedded
  Technology and Applications Symposium (RTAS)}, pp.~174--187, 2020.

\bibitem{anytimenet}
J.-E. Kim, R.~Bradford, and Z.~Shao, ``Anytimenet: Controlling time-quality
  tradeoffs in deep neural network architectures,'' in {\em 2020 Design,
  Automation Test in Europe Conference Exhibition (DATE)}, pp.~945--950, 2020.

\bibitem{apnet}
S.~Bateni and C.~Liu, ``Apnet: Approximation-aware real-time neural network,''
  in {\em 2018 IEEE Real-Time Systems Symposium (RTSS)}, pp.~67--79, 2018.

\bibitem{imprecisecomp}
S.~Yao, Y.~Hao, Y.~Zhao, H.~Shao, D.~Liu, S.~Liu, T.~Wang, J.~Li, and
  T.~Abdelzaher, ``Scheduling real-time deep learning services as imprecise
  computations,'' in {\em 2020 IEEE 26th International Conference on Embedded
  and Real-Time Computing Systems and Applications (RTCSA)}, pp.~1--10, 2020.

\bibitem{subflow}
S.~Lee and S.~Nirjon, ``Subflow: A dynamic induced-subgraph strategy toward
  real-time dnn inference and training,'' in {\em 2020 IEEE Real-Time and
  Embedded Technology and Applications Symposium (RTAS)}, pp.~15--29, 2020.

\bibitem{rttasksched}
S.~Liu, S.~Yao, X.~Fu, H.~Shao, R.~Tabish, S.~Yu, A.~Bansal, H.~Yun, L.~Sha,
  and T.~Abdelzaher, ``Real-time task scheduling for machine perception in in
  intelligent cyber-physical systems,'' {\em IEEE Transactions on Computers},
  pp.~1--1, 2021.

\bibitem{voxelrcnn}
J.~Deng, S.~Shi, P.~Li, W.~Zhou, Y.~Zhang, and H.~Li, ``Voxel {R-CNN:} towards
  high performance voxel-based 3d object detection,'' {\em CoRR},
  vol.~abs/2012.15712, 2020.

\bibitem{fasterrcnn}
S.~Ren, K.~He, R.~B. Girshick, and J.~Sun, ``Faster {R-CNN:} towards real-time
  object detection with region proposal networks,'' {\em CoRR},
  vol.~abs/1506.01497, 2015.

\bibitem{pointpillars}
A.~H. Lang, S.~Vora, H.~Caesar, L.~Zhou, J.~Yang, and O.~Beijbom,
  ``Pointpillars: Fast encoders for object detection from point clouds,'' in
  {\em 2019 IEEE/CVF Conference on Computer Vision and Pattern Recognition
  (CVPR)}, pp.~12689--12697, 2019.

\bibitem{yang2019std}
Z.~Yang, Y.~Sun, S.~Liu, X.~Shen, and J.~Jia, ``Std: Sparse-to-dense 3d object
  detector for point cloud,'' in {\em Proceedings of the IEEE/CVF International
  Conference on Computer Vision}, pp.~1951--1960, 2019.

\bibitem{zhou2020tracking}
X.~Zhou, V.~Koltun, and P.~Kr{\"a}henb{\"u}hl, ``Tracking objects as points,''
  in {\em European Conference on Computer Vision}, pp.~474--490, Springer,
  2020.

\bibitem{redmon2018yolov3}
J.~Redmon and A.~Farhadi, ``Yolov3: An incremental improvement,'' {\em arXiv
  preprint arXiv:1804.02767}, 2018.

\bibitem{jetson-agx}
{NVIDIA}, ``{Jetson AGX Xavier Developer Kit}.''
  https://developer.nvidia.com/embedded/jetson-agx-xavier-developer-kit.

\bibitem{pvrcnn}
S.~Shi, C.~Guo, L.~Jiang, Z.~Wang, J.~Shi, X.~Wang, and H.~Li, ``Pv-rcnn:
  Point-voxel feature set abstraction for 3d object detection,'' in {\em 2020
  IEEE/CVF Conference on Computer Vision and Pattern Recognition (CVPR)},
  pp.~10526--10535, 2020.

\bibitem{second}
Y.~Yan, Y.~Mao, and B.~Li, ``{SECOND}: Sparsely embedded convolutional
  detection,'' {\em Sensors}, pp.~3337--3354, 2018.

\bibitem{centerpoint}
T.~Yin, X.~Zhou, and P.~Kr{\"a}henb{\"u}hl, ``Center-based 3d object detection
  and tracking,'' {\em CVPR}, 2021.

\bibitem{parta2}
S.~Shi, Z.~Wang, J.~Shi, X.~Wang, and H.~Li, ``From points to parts: 3d object
  detection from point cloud with part-aware and part-aggregation network,''
  {\em IEEE Transactions on Pattern Analysis and Machine Intelligence},
  vol.~43, no.~8, pp.~2647--2664, 2021.

\bibitem{cadnn}
C.~Reading, A.~Harakeh, J.~Chae, and S.~L. Waslander, ``Categorical depth
  distribution network for monocular 3d object detection,'' in {\em 2021
  IEEE/CVF Conference on Computer Vision and Pattern Recognition (CVPR)},
  pp.~8551--8560, 2021.

\bibitem{megvii}
B.~Zhu, Z.~Jiang, X.~Zhou, Z.~Li, and G.~Yu, ``Class-balanced grouping and
  sampling for point cloud 3d object detection,'' {\em CoRR},
  vol.~abs/1908.09492, 2019.

\bibitem{pixor}
B.~Yang, W.~Luo, and R.~Urtasun, ``Pixor: Real-time 3d object detection from
  point clouds,'' in {\em 2018 IEEE/CVF Conference on Computer Vision and
  Pattern Recognition}, pp.~7652--7660, 2018.

\bibitem{abc}
J.-E. Kim, R.~Bradford, M.-K. Yoon, and Z.~Shao, ``Abc: Abstract prediction
  before concreteness,'' in {\em 2020 Design, Automation Test in Europe
  Conference Exhibition (DATE)}, pp.~1103--1108, 2020.

\bibitem{openpcdet2020}
O.~D. Team, ``Openpcdet: An open-source toolbox for 3d object detection from
  point clouds.'' https://github.com/open-mmlab/OpenPCDet, 2020.

\bibitem{nuscenes}
H.~Caesar, V.~Bankiti, A.~H. Lang, S.~Vora, V.~E. Liong, Q.~Xu, A.~Krishnan,
  Y.~Pan, G.~Baldan, and O.~Beijbom, ``nuscenes: A multimodal dataset for
  autonomous driving,'' in {\em 2020 IEEE/CVF Conference on Computer Vision and
  Pattern Recognition (CVPR)}, pp.~11618--11628, 2020.

\bibitem{lidaro}
Y.~Li and J.~Ibanez-Guzman, ``Lidar for autonomous driving: The principles,
  challenges, and trends for automotive lidar and perception systems,'' {\em
  IEEE Signal Processing Magazine}, vol.~37, no.~4, pp.~50--61, 2020.

\bibitem{waymo}
P.~Sun, H.~Kretzschmar, X.~Dotiwalla, A.~Chouard, V.~Patnaik, P.~Tsui, J.~Guo,
  Y.~Zhou, Y.~Chai, B.~Caine, V.~Vasudevan, W.~Han, J.~Ngiam, H.~Zhao,
  A.~Timofeev, S.~Ettinger, M.~Krivokon, A.~Gao, A.~Joshi, Y.~Zhang, J.~Shlens,
  Z.~Chen, and D.~Anguelov, ``Scalability in perception for autonomous driving:
  Waymo open dataset,'' in {\em 2020 IEEE/CVF Conference on Computer Vision and
  Pattern Recognition (CVPR)}, pp.~2443--2451, 2020.

\bibitem{lyft}
R.~Kesten, M.~Usman, J.~Houston, T.~Pandya, K.~Nadhamuni, A.~Ferreira, M.~Yuan,
  B.~Low, A.~Jain, P.~Ondruska, S.~Omari, S.~Shah, A.~Kulkarni, A.~Kazakova,
  C.~Tao, L.~Platinsky, W.~Jiang, and V.~Shet, ``Level 5 perception dataset
  2020.'' https://level-5.global/level5/data/, 2019.

\bibitem{autoware}
S.~Kato, S.~Tokunaga, Y.~Maruyama, S.~Maeda, M.~Hirabayashi, Y.~Kitsukawa,
  A.~Monrroy, T.~Ando, Y.~Fujii, and T.~Azumi, ``Autoware on board: Enabling
  autonomous vehicles with embedded systems,'' in {\em 2018 ACM/IEEE 9th
  International Conference on Cyber-Physical Systems (ICCPS)}, pp.~287--296,
  2018.

\bibitem{apollo}
``{Baidu Apollo team (2017), Apollo: Open Source Autonomous Driving}.''
  https://github.com/ApolloAuto/apollo.
\newblock Accessed: 2022-04-29.

\bibitem{rtm3d}
P.~Li, H.~Zhao, P.~Liu, and F.~Cao, ``{RTM3D:} real-time monocular 3d detection
  from object keypoints for autonomous driving,'' {\em CoRR},
  vol.~abs/2001.03343, 2020.

\bibitem{complexyolo}
M.~Simon, S.~Milz, K.~Amende, and H.~Gross, ``Complex-yolo: Real-time 3d object
  detection on point clouds,'' {\em CoRR}, vol.~abs/1803.06199, 2018.

\bibitem{deepcompression}
S.~Han, H.~Mao, and W.~J. Dally, ``Deep compression: Compressing deep neural
  network with pruning, trained quantization and huffman coding,'' in {\em 4th
  International Conference on Learning Representations, {ICLR} 2016, San Juan,
  Puerto Rico, May 2-4, 2016, Conference Track Proceedings} (Y.~Bengio and
  Y.~LeCun, eds.), 2016.

\bibitem{deepiot}
S.~Yao, Y.~Zhao, A.~Zhang, L.~Su, and T.~F. Abdelzaher, ``Compressing deep
  neural network structures for sensing systems with a compressor-critic
  framework,'' {\em CoRR}, vol.~abs/1706.01215, 2017.

\bibitem{fastdeepiot}
S.~Yao, Y.~Zhao, H.~Shao, S.~Liu, D.~Liu, L.~Su, and T.~F. Abdelzaher,
  ``Fastdeepiot: Towards understanding and optimizing neural network execution
  time on mobile and embedded devices,'' in {\em Proceedings of the 16th {ACM}
  Conference on Embedded Networked Sensor Systems, SenSys 2018, Shenzhen,
  China, November 4-7, 2018} (G.~S. Ramachandran and B.~Krishnamachari, eds.),
  pp.~278--291, {ACM}, 2018.

\bibitem{adaptivequant}
Y.~Zhou, S.~Moosavi{-}Dezfooli, N.~Cheung, and P.~Frossard, ``Adaptive
  quantization for deep neural network,'' {\em CoRR}, vol.~abs/1712.01048,
  2017.

\bibitem{cascaded}
B.~Minnehan and A.~E. Savakis, ``Cascaded projection: End-to-end network
  compression and acceleration,'' {\em CoRR}, vol.~abs/1903.04988, 2019.

\bibitem{hu2021exploring}
Y.~Hu, S.~Liu, T.~Abdelzaher, M.~Wigness, and P.~David, ``On exploring image
  resizing for optimizing criticality-based machine perception,'' in {\em 2021
  IEEE 27th International Conference on Embedded and Real-Time Computing
  Systems and Applications (RTCSA)}, pp.~169--178, IEEE, 2021.

\bibitem{liu2020removing}
S.~Liu, S.~Yao, X.~Fu, R.~Tabish, S.~Yu, A.~Bansal, H.~Yun, L.~Sha, and
  T.~Abdelzaher, ``On removing algorithmic priority inversion from
  mission-critical machine inference pipelines,'' in {\em 2020 IEEE Real-Time
  Systems Symposium (RTSS)}, pp.~319--332, IEEE, 2020.

\end{thebibliography}

\end{document}